# Exploring Thematic Coherence in Fake News

Martins Samuel Dogo, Deepak P and Anna Jurek-Loughrey

Queen's University Belfast, UK
mdogo01@qub.ac.uk, deepaksp@acm.org, a.jurek@qub.ac.uk

**Abstract.** The spread of fake news remains a serious global issue; understanding and curtailing it is paramount. One way of differentiating between deceptive and truthful stories is by analyzing their coherence. This study explores the use of topic models to analyze the coherence of cross-domain news shared online. Experimental results on seven cross-domain datasets demonstrate that fake news shows a greater thematic deviation between its opening sentences and its remainder.

## 1 Introduction

The impact of news on our daily affairs is greater than it has ever been. Fabrication and dissemination of falsehood have become politically lucrative endeavors, thereby harming public discourse and worsening political polarization [1]. These motivations have led to a complex and continuously evolving phenomenon mainly characterized by dis- and misinformation, commonly collectively referred to as fake news [2]. This denotes various kinds of false or unverified information, which may vary based on their authenticity, intention, and format [3]. Shu et al. [4] define it as "a news article that is intentionally and verifiably false."

The dissemination of fake news is increasing, and because it appears in various forms and self-reinforces [1, 3], it is difficult to erode. Therefore, there is an urgent need for increased research in understanding and curbing it. This paper considers fake news that appears in the form of long online articles and explores the extent of internal consistency within fake news vis-à-vis legitimate news. In particular, we run experiments to determine whether thematic deviations — i.e., a measure of how dissimilar topics discussed in different parts of an article are — between the opening and remainder sections of texts can be used to distinguish between fake and real news across different news domains.

### 1.1 Motivation

A recent study suggests that some readers may skim through an article instead of reading the whole content because they overestimate their political knowledge, while others may hastily share news without reading it fully, for emotional affirmation [5].

This presents bad actors with the opportunity of deftly interspersing news content with falsity. Moreover, the production of fake news typically involves the collation of disjoint content and lacks a thorough editorial process [6].

Topics discussed in news pieces can be studied to ascertain whether the article thematically deviates between its opening and the rest of the story, or if it remains coherent throughout. Thematic analysis is useful here for two reasons. First, previous studies show that the coherence between units of discourse (such as sentences) in a document is useful for determining its veracity [6, 7]. Second, analysis of thematic deviation can identify general characteristics of fake news that persist across multiple news domains.

Although topics have been employed as features [8–10], they have not been applied to study the unique characteristics of fake news. Research efforts in detecting fake news through thematic deviation have thus far focused on spotting incongruences between pairs of headlines and body texts [11–14]. Yet, thematic deviation can also exist within the body text of a news item. Our focus is to examine these deviations to distinguish fake from real news.

To the best of the authors' knowledge, this is the first work that explores thematic deviations in the body text of news articles to distinguish between fake and legitimate news.

## 2 Related Work

The coherence of a story may be indicative of its veracity. For example, [7] demonstrated this by applying Rhetorical Structure Theory [15] to study the discourse of deceptive stories posted online. They found that a major distinguishing characteristic of deceptive stories is that they are disjunctive. Also, while truthful stories provide evidence and restate information, deceptive ones do not. This suggests that false stories may tend to thematically deviate more due to disjunction, while truthful stories are likely to be more coherent due to restatement. Similarly, [6] investigated the coherence of fake and real news by learning hierarchical structures based on sentence-level dependency parsing. Their findings also suggest that fake news documents are less coherent.

Topic models are unsupervised algorithms that aid the identification of themes discussed in large corpora. One example is Latent Dirichlet Allocation (LDA), which is a generative probabilistic model that aids the discovery of latent themes or topics in a corpus [16]. Vosoughi et al. [17] used LDA to show that false rumor tweets tend to be more novel than true ones. Novelty was evaluated using three measures: Information Uniqueness, Bhattacharyya Distance, and Kullback-Leibler Divergence. Likewise, [18] used LDA to assess the credibility of Twitter users by analyzing the topical divergence of their tweets to those of other users. They also assessed the veracity of users' tweets by comparing the topic distributions of new tweets against historically discussed topics. Divergence was computed using the Jensen-Shannon Divergence, Root Mean Squared Error, and Squared Error. Our work primarily differs from these two in that we analyze full-length articles instead of tweets.

## 3   Research Goal and Contributions

This research aims to assess the importance of internal consistency within articles as a high-level feature to distinguish between fake and real news stories across different domains. We set out to explore whether the opening segments of fake news thematically deviates from the rest of it, significantly more than in authentic news. We experiment with seven datasets which collectively cover a wide variety of news domains, from business to celebrity, to warfare. Deviations are evaluated by calculating the distance between the topic distribution of the opening part of an article, to that of its remainder. We take the first five sentences of an article as its opening segment.

Our contributions are summarized as follows:

1. We present new insights towards understanding the underlying characteristics of fake news, based on thematic deviations between the opening and remainder parts of news body text.
2. We carry out experiments on five cross-domain datasets. The results demonstrate the effectiveness of thematic deviation for distinguishing fake from real news.

## 4   Experiments

We hypothesize the following: *the opening sentences of a false news article will tend to thematically deviate more from the rest of it, as compared to an authentic article*. To test this hypothesis, we carried out experiments in the manner shown in Algorithm 1. We use Python and open-source packages for all computations.

**Procedure.** All articles ($S_{bg}$) are split into two parts: its first $x$ [1] sentences, and the remaining $y$. Next, $N$ topics are obtained from $x$ and $y$ using an LDA model trained using Gensim[2] on the entire dataset. For $i = (1, \ldots, m)$ topics, let $p_x = (p_{x1}, \ldots, p_{xm})$ and $p_y = (p_{y1}, \ldots, p_{ym})$ be two vectors of topic distributions, which denote the prevalence of a topic $i$ in the opening text $x$ and remainder $y$ of an article, respectively. The following are metrics used to measure the topical divergence between parts $x$ and $y$ of an article:

- Chebyshev ($D_{Ch}$):

$$D_{Ch}(p_{xi}, p_{yi}) = \max_i |p_{xi} - p_{yi}| \qquad (1)$$

- Euclidean ($D_E$):

$$D_E(p_{xi}, p_{yi}) = \|p_{xi} - p_{yi}\| = \sqrt{\sum_{i=1}^{m}(p_{xi} - p_{yi})^2} \qquad (2)$$

---

[1] We used only articles with at least $x + 1$ sentences.
[2] https://radimrehurek.com/gensim/

4- Squared Euclidean ($D_{SE}$):

$$D_{SE}(p_{xi}, p_{yi}) = \sum_{i=1}^{m}(p_{xi} - p_{yi})^2 \quad (3)$$

Intuitively, Chebyshev distance is the greatest difference found between any two topics in $x$ and $y$. The Euclidean distance measures how "far" the two topic distributions are from one another, while the Squared Euclidean distance is simply the square of that "farness".

Finally, the average and median values of each distance are calculated across all fake ($S_f$) and real ($S_r$) articles. We repeated these steps with varying values of $N$ (from 10 to 200 topics) and $x$ (from 1 to 5 sentences).

**Algorithm 1.** Corpora comparison procedure.

---

**Data:** A corpus $S_{bg} = S_f \cup S_r$ of full-length fake ($S_f = \{d_1^f, d_2^f, ..., d_F^f\}$) and real ($S_r = \{d_1^r, d_2^r, ..., d_R^r\}$) documents

**Input:** Pairs of first $l = [1, 2, ..., 5]$ sentences and remainder $y$ of each fake ($d_i^f = \langle d_{i_x}^f, d_{i_y}^f \rangle$; $|d_{i_x}^f| = l$) and real article ($d_i^r = \langle d_{i_x}^r, d_{i_y}^r \rangle$; $|d_{i_x}^r| = l$); LDA model $\mathcal{M}_{bg}$ generated using $S_{bg}$; the number of topics $N \in \{10, 20, 50, 100, 150, 200\}$; divergence functions $\mathcal{D} \in \{D_{Ch}, D_E, D_{SE}\}$;

1:  **foreach** $l = [1, 2, ..., 5]$ **do**
2:    **foreach** article $\langle d_{i_x}^f, d_{i_y}^f \rangle$ **do**
3:      get the distribution of $N$ topics in $d_{i_x}^f$ and $d_{i_y}^f$ using $\mathcal{M}_{bg}$
4:      $T_{i_x}^f = (p_i^x, ..., p_N^x); T_{i_y}^f = (p_i^y, ..., p_N^y)$
5:      get the distribution of $N$ topics in $d_{i_x}^r$ and $d_{i_y}^r$ using $\mathcal{M}_{bg}$
6:      $T_{i_x}^r = (p_i^x, ..., p_N^x); T_{i_y}^r = (p_i^y, ..., p_N^y)$
7:      $D_i^f = \mathcal{D}(T_{i_x}^f, T_{i_y}^f); D_i^r \mathcal{D}(T_{i_x}^r, T_{i_y}^r)$
8:    **end**
9:    $D_{avg}^f = \text{average}(D_i^f \mid i \in \{1, ..., F\}); D_{avg}^r = \text{average}(D_i^r \mid i \in \{1, ..., R\})$
10:   $D_{med}^f = \text{median}(D_i^f \mid i \in \{1, ..., F\}); D_{med}^r = \text{median}(D_i^r \mid i \in \{1, ..., R\})$
11:   **return** $\{D_{avg}^f, D_{avg}^r, D_{med}^f, D_{med}^r\}$
12: **end**

---

**Pre-processing.** Articles are split into sentences using the NTLK[3] package. Each sentence is tokenized and lowercased to form a list of words, from which stop words are removed. Bigrams are then formed and added to the vocabulary. Next, each document is lemmatized using spaCy[4], and only noun, adjective, verb, and adverb lemmas are retained. A dictionary is formed by applying these steps to $S_{bg}$. Each document is converted into a bag-of-words (BoW) format, which is used to create an LDA model ($\mathcal{M}_{bg}$). Fake and real articles are subsequently pre-processed likewise (i.e., from raw text data to BoW format) before topics are extracted from them.

---

[3] https://www.nltk.org/
[4] https://spacy.io/models/en/

We consider the opening sentences of articles to be sufficient for capturing the lead or "opening theme" of the story, which will likely fall within the first paragraph. The first paragraph may in some cases be either too short or long for this, especially for fake articles that often lack a proper structure. Overly short or lengthy texts will influence the extraction of topics more adversely than if a set number of sentences are used.

**Datasets.** Table 1 summarizes the datasets (after pre-processing) used in this study and lists the domains (as stated by the dataset provider) covered by each. An article's length (Avg. length) is measured by the number of words that remain after pre-processing. The article maximum lengths (Max. length) is measured in terms of the number of sentences. We use the following datasets:

- BuzzFeed-Webis Fake News Corpus 2016[5] (BuzzFeed-Web) [19]
- BuzzFeed Political News Data[6] (BuzzFeed-Political) [20]
- FakeNewsAMT + Celebrity (AMT+C) [21]
- Falsified and Legitimate Political News Database[7] (POLIT)
- George McIntire's fake news dataset (GMI)[8]
- University of Victoria's Information Security and Object Technology (ISOT)[9] Research Lab [22]
- Syrian Violations Documentation Centre (SVDC)[10] [23]

**Table 1.** Summary of datasets after pre-processing (f – fake, r – real).

| Dataset (domain) | No. of fake | No. of real | Avg. length of sentences in words (f) | Avg. length of sentences in words (r) | Max. length (f) | Max. length (r) |
|---|---|---|---|---|---|---|
| AMT+C (business, education, entertainment, politics, sports, tech) | 324 | 317 | 14.7 | 23.2 | 64 | 1,059 |
| BuzzFeed-Political (politics) | 116 | 127 | 18.9 | 43.9 | 76 | 333 |
| BuzzFeed-Web (politics) | 331 | 1,214 | 21.7 | 26.4 | 117 | 211 |

---

[5] https://zenodo.org/record/1239675
[6] https://github.com/BenjaminDHorne/fakenewsdata1
[7] http://victoriarubin.fims.uwo.ca/news-verification/access-polit-false-n-legit-news-db-2016-2017/
[8] https://github.com/GeorgeMcIntire/fake_real_news_dataset (accessed 5 November 2018)
[9] https://www.uvic.ca/engineering/ece/isot/
[10] https://zenodo.org/record/2532642

| Dataset (domain) | No. of fake | No. of real | Avg. length of sentences in words (f) | Avg. length of sentences in words (r) | Max. length (f) | Max. length (r) |
|---|---|---|---|---|---|---|
| GMI (politics) | 2,695 | 2,852 | 33.9 | 42.8 | 1,344 | 406 |
| ISOT (government, politics) | 19,324 | 16,823 | 18.0 | 20.3 | 289 | 324 |
| POLIT (politics) | 122 | 134 | 19.2 | 34.9 | 96 | 210 |
| SVDC (conflict, war) | 312 | 352 | 14.0 | 14.6 | 62 | 49 |

**Evaluation.** We evaluate differences in coherence of fake and real articles using the T-test at 5% significance level. The null hypothesis is that the mean coherence of fake and real news is equal. The alternative hypothesis is that the mean coherence of real news is greater than that of fake news. We expect that there will be a greater topic deviation in fake news and thus, its coherence will be lesser than that of real news.

## 5    Results and Discussion

Results of the experimental evaluation using the different divergence measures are shown in Table 2. We observe that fake news is generally likely to show greater thematic deviation (lesser coherence) than real news in all datasets. Table 3 shows the mean $D_{Ch}$ deviations of fake and real articles across $N=\{10, 20, 30, 40, 50, 100, 150, 200\}$ topics. Although results for AMT+C and BuzzFeed-Web are not statistically significant according to the T-test and therefore, do not meet our expectations, results for all other datasets are. Nonetheless, the mean and median values for fake news are lower than those of real news for these datasets. Table 3, which shows mean and median $D_{Ch}$ deviations of fake and real articles across all values of $N$. Fig. 1 shows mean and median results for comparing topics in the first five and the remaining sentences. Results for values of $N$ not shown are similar (with $D_{Ch}$ gradually decreasing as $N$ increases).

**Table 2.** Results of T-test evaluation based on different measures of deviation used.

| Dataset | p-value ($D_{Ch}$) | p-value ($D_E$) | p-value ($D_{SE}$) |
|---|---|---|---|
| AMT+C | 0.144 | 0.126 | 0.116 |
| BuzzFeed-Political | 0.0450 | 0.0147 | 0.0287 |
| BuzzFeed-Web | 0.209 | 0.209 | 0.207 |
| GMI | 0.0480 | 0.00535 | 0.0106 |
| ISOT | 0.00319 | 0.000490 | 0.000727 |
| POLIT | 0.000660 | 0.0000792 | 0.0000664 |
| SVDC | 0.000684 | 0.0000112 | 0.0000789 |

**Table 3.** Mean and median $D_{Ch}$ deviations of $N=\{10, 20, 30, 40, 50, 100, 150, 200\}$ topics combined for fake and real news (f – fake, r – real).

| Dataset | Mean $D_{Ch}$(f) | Mean $D_{Ch}$(r) | Median $D_{Ch}$(f) | Median $D_{Ch}$(r) |
| --- | --- | --- | --- | --- |
| AMT+C | 0.2568 | 0.2379 | 0.2438 | 0.2285 |
| BuzzFeed-Political | 0.2373 | 0.2149 | 0.2345 | 0.2068 |
| BuzzFeed-Web | 0.2966 | 0.2812 | 0.2863 | 0.2637 |
| GMI | 0.4580 | 0.4241 | 0.4579 | 0.4222 |
| ISOT | 0.3372 | 0.2971 | 0.3369 | 0.2989 |
| POLIT | 0.2439 | 0.1939 | 0.2416 | 0.1894 |
| SVDC | 0.2975 | 0.2517 | 0.2934 | 0.2435 |

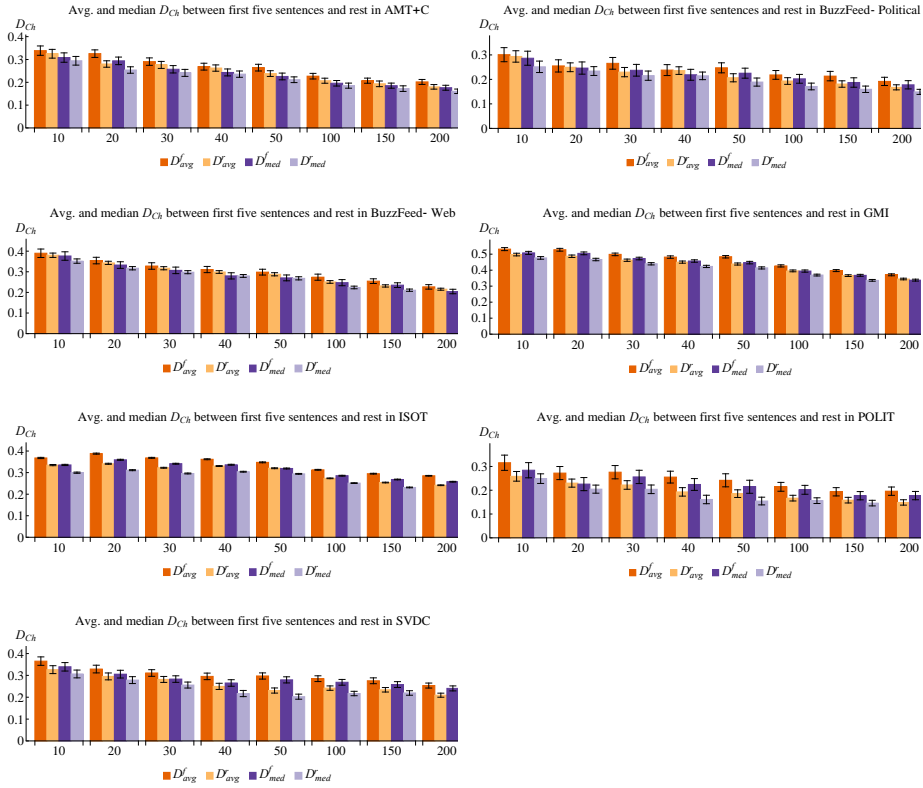

**Fig. 1.** Average and median Chebyshev distances in fake and real news, when comparing topics in the first five sentences to the rest of each article. Error bars show 95% confidence interval.

We found that comparing the first five sentences to the rest of the article yielded the best results (i.e., greatest disparity between fake and real deviations) for most datasets and measures. This is likely due to the first five sentences containing more information. For example, five successive sentences are likely to entail one another and contribute more towards a topic than a single sentence.

It is worth highlighting the diversity of datasets used here, in terms of domain, size, and the nature of articles. For example, the fake and real news in the SVDC dataset have a very similar structure. Both types of news were mostly written with the motivation to inform the reader on conflict-related events that took across Syria. However, fake articles are labeled as such primarily because the reportage (e.g., on locations and number of casualties) in them is insufficiently accurate.

To gain insight into possible causes of greater deviation in fake news, we qualitatively inspected the five most and least diverging fake and real articles (according to $D_{Ch}$). We also compared a small set of low and high number of topics ($N \leq 30$ and $N \geq 100$). We observed that fake openings tend to be shorter, vaguer, and less congruent with the rest of the text. By contrast, real news openings generally give a better narrative background to the rest of the story.

Although the writing style in fake news is sometimes unprofessional, this is an unlikely reason for the higher deviations in fake news. Both fake and real news try to expand on the opening lead of the story, with more context and explanation. Indeed, we observed that real news tends to have longer sentences, which give more detailed information about a story, and are more narrative. It can be argued that the reason behind this is that fake articles are designed to get readers' attention, whereas legitimate ones are written to inform the reader. For instance, social media posts which include a link to an article are sometimes displayed with a short snippet of the article's opening text or its summary. This section can be designed to capture readers' attention.

It also conceivable that a bigger team of people working to produce a fake piece may contribute to its vagueness. They may input different perspectives that diversify the story and makes it less coherent. This may be compared to real news, whereby there typically are one or two professional writers and therefore, better coherence.

## 6    Conclusion

Fake news and deceptive stories tend to open with sentences which may be incoherent with the rest of the text. It is worth exploring if the consistency of fake and real news can distinguish between the two. Accordingly, we investigated the thematic deviations of seven cross-domain fake and real news using topic modeling. Our findings suggest that the opening sentences of fake articles topically deviate more from the rest of the article, as compared to real news. The next step is to find possible reasons behind these deviations through in-depth analyses of topics. In conclusion, this paper presents valuable insights into thematic differences between fake and authentic news, which may be exploited for fake news detection.

**References**

[1] A. E. Waldman, "The Marketplace of Fake News," *Univ. Pennsylvania J. Const. Law*, vol. 20, pp. 846–869, 2018, DOI: 10.4135/9781604265774.n911.

[2] C. Wardle, "Fake News. It's Complicated," *First Draft*, 2017. https://firstdraftnews.org/fake-news-complicated/, last accessed 2019/1/24.

[3] X. Zhou and R. Zafarani, "Fake News: A Survey of Research, Detection Methods, and Opportunities," 2018, Accessed: Nov. 10, 2019. [Online]. Available: http://arxiv.org/abs/1812.00315.

[4] K. Shu, A. Sliva, S. Wang, J. Tang, and H. Liu, "Fake News Detection on Social Media," *ACM SIGKDD Explor. Newsl.*, vol. 19, no. 1, pp. 22–36, 2017, doi: 10.1145/3137597.3137600.

[5] N. M. Anspach, J. T. Jennings, and K. Arceneaux, "A little bit of knowledge: Facebook's News Feed and self-perceptions of knowledge," *Res. Polit.*, vol. 6, no. 1, 2019, DOI: 10.1177/2053168018816189.

[6] H. Karimi and J. Tang, "Learning Hierarchical Discourse-level Structure for Fake News Detection," in *Proceedings of the 2019 Conference of the North*, 2019, pp. 3432–3442, DOI: 10.18653/v1/N19-1347.

[7] V. L. Rubin and T. Lukoianova, "Truth and deception at the rhetorical structure level," *J. Assoc. Inf. Sci. Technol.*, vol. 66, no. 5, pp. 905–917, May 2015, doi: 10.1002/asi.23216.

[8] S. Das Bhattacharjee, A. Talukder, and B. V. Balantrapu, "Active learning based news veracity detection with feature weighting and deep-shallow fusion," in *Proceedings - 2017 IEEE International Conference on Big Data, Big Data 2017*, Dec. 2018, vol. Jan. 2018, pp. 556–565, DOI: 10.1109/BigData.2017.8257971.

[9] A. Benamira, B. Devillers, E. Lesot, A. K. Ray, M. Saadi, and F. D. Malliaros, "Semi-Supervised Learning and Graph Neural Networks for Fake News Detection," pp. 568–569, Aug. 2019, Accessed: Nov. 29, 2019. [Online]. Available: https://hal.archives-ouvertes.fr/hal-02334445/.

[10] S. Li, K. Ma, X. Niu, Y. Wang, K. Ji, Z. Yu and Z. Chen, "Stacking-based ensemble learning on low dimensional features for fake news detection," in *Proceedings - 17th IEEE International Conference on Smart City and 5th IEEE International Conference on Data Science and Systems, HPCC/SmartCity/DSS 2019*, Aug. 2019, pp. 2730–2735, DOI: 10.1109/HPCC/SmartCity/DSS.2019.00383.

[11] Y. Chen, N. J. Conroy, and V. L. Rubin, "Misleading online content: Recognizing clickbait as 'false news,'" in *WMDD 2015 - Proceedings of the ACM Workshop on Multimodal Deception Detection, co-located with ICMI 2015*, 2015, pp. 15–19, DOI: 10.1145/2823465.2823467.

[12] D. S. Sisodia, "Ensemble learning approach for clickbait detection using article headline features," *Informing Sci.*, vol. 22, no. 2019, pp. 31–44, 2019, doi: 10.28945/4279.

[13] W. Ferreira and A. Vlachos, "Emergent: A novel data-set for stance classification," in *2016 Conference of the North American Chapter of the*

*Association for Computational Linguistics: Human Language Technologies, NAACL HLT 2016 - Proceedings of the Conference*, 2016, pp. 1163–1168, DOI: 10.18653/v1/n16-1138.

[14] S. Yoon, K. Park, J. Shin, H. Lim, S. Won, M. Cha and K. Jung, "Detecting Incongruity between News Headline and Body Text via a Deep Hierarchical Encoder," *Proc. AAAI Conf. Artif. Intell.*, vol. 33, pp. 791–800, 2019, doi: 10.1609/aaai.v33i01.3301791.

[15] W. C. Mann and S. A. Thompson, "Rhetorical Structure Theory: Toward a functional theory of text organization," *Text - Interdisciplinary Journal for the the Study of Discourse*, vol. 8, no. 3, pp. 243–281, 1988, doi: 10.1515/text.1.1988.8.3.243.

[16] D. M. Blei, A. Y. Ng, and M. I. Jordan, "Latent Dirichlet Allocation," *J. Mach. Learn. Res.*, vol. 3, no. Jan, pp. 993–1022, 2003, Accessed: Dec. 01, 2019. [Online]. Available: http://jmlr.csail.mit.edu/papers/v3/blei03a.html.

[17] S. Vosoughi, D. Roy, and S. Aral, "The spread of true and false news online," *Science*, vol. 359, no. 6380, pp. 1146–1151, Mar. 2018, DOI: 10.1126/science.aap9559.

[18] J. Ito, H. Toda, Y. Koike, J. Song, and S. Oyama, "Assessment of tweet credibility with LDA features," in *WWW 2015 Companion - Proceedings of the 24th International Conference on World Wide Web*, 2015, pp. 953–958, DOI: 10.1145/2740908.2742569.

[19] M. Potthast, J. Kiesel, K. Reinartz, J. Bevendorff, and B. Stein, "A stylometric inquiry into hyperpartisan and fake news," in *ACL 2018 - 56th Annual Meeting of the Association for Computational Linguistics, Proceedings of the Conference (Long Papers)*, 2018, vol. 1, pp. 231–240, doi: 10.18653/v1/p18-1022.

[20] B. D. Horne and S. Adali, "This Just In: Fake News Packs a Lot in Title, Uses Simpler, Repetitive Content in Text Body, More Similar to Satire than Real News," 2017, [Online]. Available: http://arxiv.org/abs/1703.09398.

[21] V. Pérez-Rosas, B. Kleinberg, A. Lefevre, and R. Mihalcea, "Automatic Detection of Fake News," in *Proceedings of the 27th International Conference on Computational Linguistics*, 2018, pp. 3391–3401, Accessed: Dec. 07, 2019. [Online]. Available: https://www.aclweb.org/anthology/C18-1287/.

[22] H. Ahmed, I. Traore, and S. Saad, "Detection of Online Fake News Using N-Gram Analysis and Machine Learning Techniques," in *Lecture Notes in Computer Science (including subseries Lecture Notes in Artificial Intelligence and Lecture Notes in Bioinformatics)*, 2017, DOI: 10.1007/978-3-319-69155-8_9.

[23] F. Abu Salem, R. Al Feel, S. Elbassuoni, M. Jaber, and M. Farah, "Dataset for fake news and articles detection," Jan. 2019, DOI: 10.5281/ZENODO.2532642.